\def\year{2022}\relax
\PassOptionsToPackage{table, prologue}{xcolor}

\documentclass[letterpaper]{article} 
\usepackage{aaai24} 
\usepackage{times}  
\usepackage{helvet}  
\usepackage{courier}  
\usepackage[hyphens]{url}  
\usepackage{graphicx} 
\def\UrlFont{\rm}  
\usepackage{natbib}  
\usepackage{caption} 

\usepackage{xcolor}

\DeclareCaptionStyle{ruled}{labelfont=normalfont,labelsep=colon,strut=off} 
\usepackage{algorithm}
\usepackage{algorithmic}
\usepackage{amsmath}
\usepackage{booktabs}
\usepackage{soul} 
\usepackage{color, xcolor}
\usepackage{xspace}
\usepackage{diagbox}
\usepackage{subcaption}
\usepackage{multirow}
\usepackage{siunitx}

\usepackage[T1]{fontenc}

\definecolor{bluegreen}{RGB}{0,128,128}

\usepackage{makecell}
\usepackage{tikz}

\usetikzlibrary{arrows.meta,
                chains,
                positioning,
                shapes.geometric
                }
                
\usepackage{newfloat}
\usepackage{listings}
\floatstyle{ruled}
\newfloat{listing}{tb}{lst}{}
\floatname{listing}{Listing}
\title{
A Multilingual Similarity Dataset for News Article Frame
}

\author {
    Xi Chen\textsuperscript{\rm 1}, 
    Mattia Samory\textsuperscript{\rm 2}, 
    Scott Hale\textsuperscript{\rm 3}, 
    David Jurgens\textsuperscript{\rm 4}, 
    Przemyslaw A. Grabowicz\textsuperscript{\rm 1}
}
\affiliations {
    \textsuperscript{\rm 1}University of Massachusetts Amherst\\
    \textsuperscript{\rm 2}Sapienza University of Rome\\
    \textsuperscript{\rm 3}University of Oxford\\
    \textsuperscript{\rm 4}University of Michigan\\
    xchen4@umass.edu, mattia.samory@uniroma1.it,
    scott.hale@oii.ox.ac.uk, jurgens@umich.edu, grabowicz@cs.umass.edu
}

\begin{document}

\maketitle

\begin{abstract}

Understanding the writing frame of news articles is vital for addressing social issues, and thus has attracted notable attention in the fields of communication studies. Yet, assessing such news article frames remains a challenge due to the absence of a concrete and unified standard dataset that considers the comprehensive nuances within news content.


To address this gap, we introduce an extended version of a large labeled news article dataset with 16,687 new labeled pairs. 
Leveraging the pairwise comparison of news articles, our method frees the work of manual identification of frame classes in traditional news frame analysis studies. Overall we introduce the most extensive cross-lingual news article similarity dataset available to date with 26,555 labeled news article pairs across 10 languages. Each data point has been meticulously annotated according to a codebook detailing eight critical aspects of news content, under a human-in-the-loop framework. Application examples demonstrate its potential in unearthing country communities within global news coverage, exposing media bias among news outlets, and quantifying the factors related to news creation. We envision that this news similarity dataset will broaden our understanding of the media ecosystem in terms of news coverage of events and perspectives across countries, locations, languages, and other social constructs. By doing so, it can catalyze advancements in social science research and applied methodologies, thereby exerting a profound impact on our society.

\end{abstract}

\section*{Introduction}

Every day, the world's media landscape is enriched with hundreds of thousands of news articles, spanning a multitude of languages and emanating from various corners of the globe. The ability to discern which articles narrate the same story is not just pivotal for refining news aggregation applications but also serves as a gateway to cross-linguistic analysis of media consumption and attention patterns. However, the task of measuring story congruence within these articles is fraught with complexities. Diverse dimensions in storytelling mean that even articles with substantial textual similarities may diverge significantly, recounting similar events that transpired years apart.

In the realm of communication studies, two long-term or cognitively-driven media effects stand out: the agenda-setting theory and the framing theory -- \citet{fourie2001media} defined cognition as our capacity for understanding and interpreting information in a specific way, which in turn shapes our behavior and thought processes. While the primary level of agenda-setting focuses on `what' a story conveys, the framing theory delves deeper into `how' the story is presented. This nuanced differentiation between `what' and `how' in news narratives is not just a linguistic or stylistic concern but fundamentally alters the impact and reception of news among its audiences.

Therefore, to comprehend if two news articles cover the same story in the same way, it is often not enough to just delve into specific facets of the events portrayed. Also, it requires understanding and evaluating the way to present the events, such as the writing frame.



Although frame theory has gained widespread recognition as a vital area in media research, sparking extensive study and debate, a consensus on a common and unified definition of `frame' remains missing. A traditional approach involves categorizing news articles into specific categories of frames. However, this method often restricts articles to a limited number of coarse-grained 
themes that are predefined by researchers, such as crises \cite{an2009news}, gun violence \cite{liu2019detecting, akyurek2020multi}, or policy issues \cite{card2015media}. Such limitations inhibit the potential for these methodologies to be broadly generalized.

\begin{figure}[t]
\centering
\includegraphics[width=0.98\columnwidth]{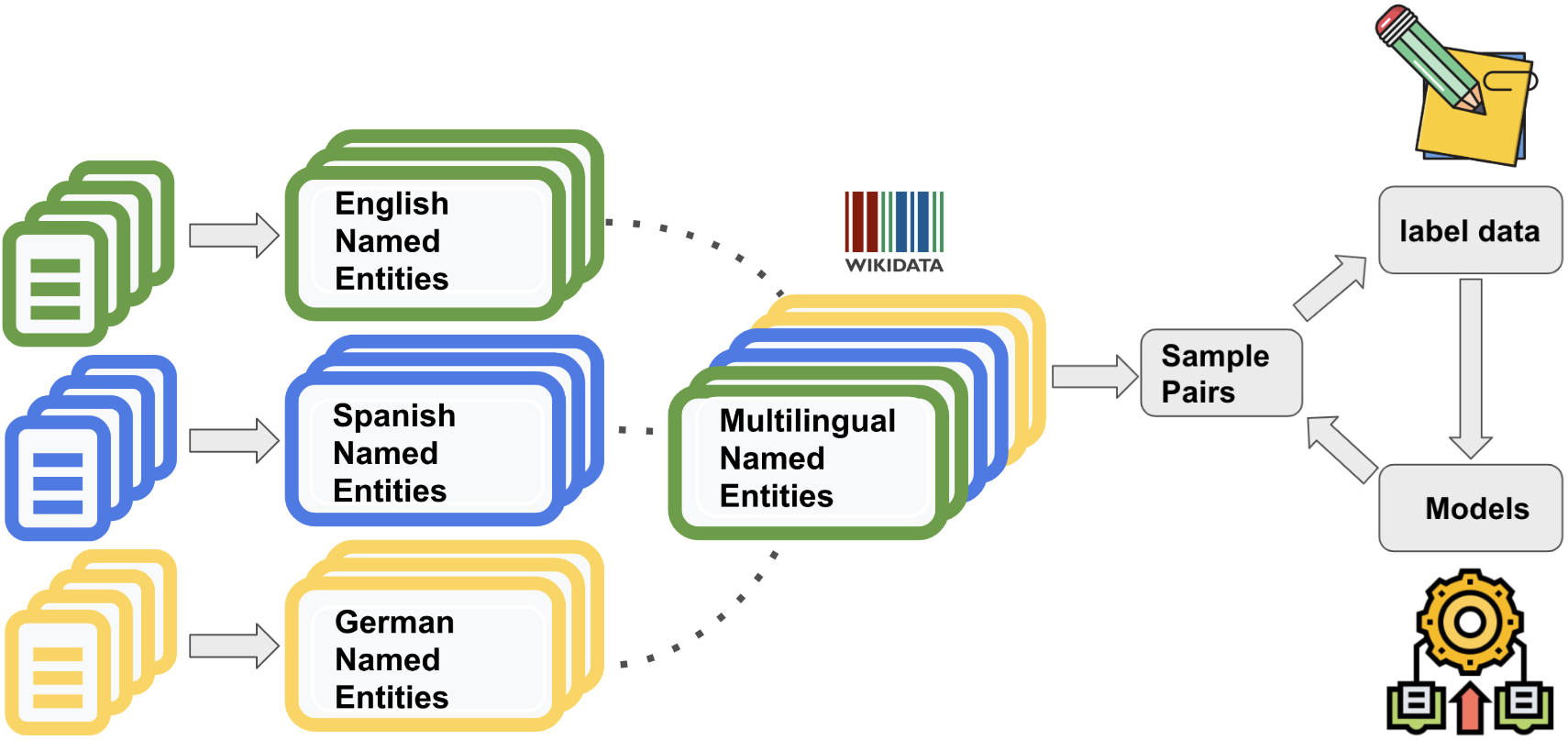}
\caption{Illustration of sample selection pipeline for annotation.\protect\footnotemark}
\label{fig:data_pipe}
\end{figure}

\footnotetext{Images by Vector Stall and Eucalyp from Flaticon.com.}


Additionally, most studies in this domain focus on sentence-level analysis, constrained by data availability and suitable large-scale analytical approaches. This results in studies that often examine only news headlines \cite{liu2019detecting, akyurek2020multi} or paragraph contexts \cite{card2015media}, thereby failing to capture the comprehensive essence of the news articles.


\begin{figure}[t!]
    \centering
\noindent\fbox{%
\parbox{0.46\textwidth}{%
\centering
\vspace{-2mm}    
\begin{description} 
\scriptsize
\itemsep0em 
\item[GEO] How similar is the geographic focus (places, cities, countries, etc.) of the two articles?
\item[ENT] How similar are the named entities (e.g., people, companies, organizations, products, named living beings), excluding previously considered locations appearing in the two articles?
\item[TIME] Are the two articles relevant to similar time periods or describing similar time periods?
\item[NAR] How similar are the narrative schemas presented in the two articles?
\item[OVERALL] Overall, are the two articles covering the same substantive news story? (excluding style, framing, and tone)
\item[STYLE] Do the articles have similar writing styles?
\item[TONE] Do the articles have similar tones?
\item[\textit{FRAME}] \textit{Do the articles have similar framing and express similar opinions?}
\end{description}
\vspace{-2mm}
}
}
    \caption{The annotation scheme used by \citet{chen2022semeval}, which we extended with the FRAME aspect (bottom line).}
    \label{fig:annotation-scheme}
\end{figure}

A relevant research area, namely targeted sentiment analysis, involves identifying named entities discussed in a document and classifying the sentiment towards them. Similar to frame theory, datasets in targeted sentiment analysis are typically limited in size and scope, focusing mainly on sentence-level data. This limitation hinders accuracy due to the absence of co-reference and discourse context, let alone fully extracting complex relationships within a document's entirety \cite{steinberger-etal-2017-large, luo2022challenges}.


In our work, we have expanded a news article dataset, effectively overcoming these shortcomings in both frame theory and targeted sentiment analysis \cite{chen2022semeval}. By adopting a human-in-the-loop framework for annotation, we ensured the creation of high-quality, large-scale data. We define the measurement of frame in news articles by expanding upon the aspects of pairwise news similarity, thus circumventing the need for narrowly defined, subjective frame categories, which are often abstract and subtle to operationalize.

Concretely we introduce the most extensive multilingual news article similarity dataset to date, containing nearly 27 thousand news article pairs across 10 languages. This dataset offers a nuanced, pairwise measure, bridging the gap between high-level per-outlet bias and low-level, per-sentence targeted sentiment and frame connotations. We believe it will establish a new benchmark for tasks like cross-lingual document matching, news clustering, and multilingual information retrieval. Furthermore, it paves the way for exploring potential media biases and agendas within different linguistic communities. Our ultimate goal is to bridge societal barriers, enhancing effective news communication and understanding in our evermore connected global society.


\begin{figure}[t]
\centering
\includegraphics[width=0.98\columnwidth]{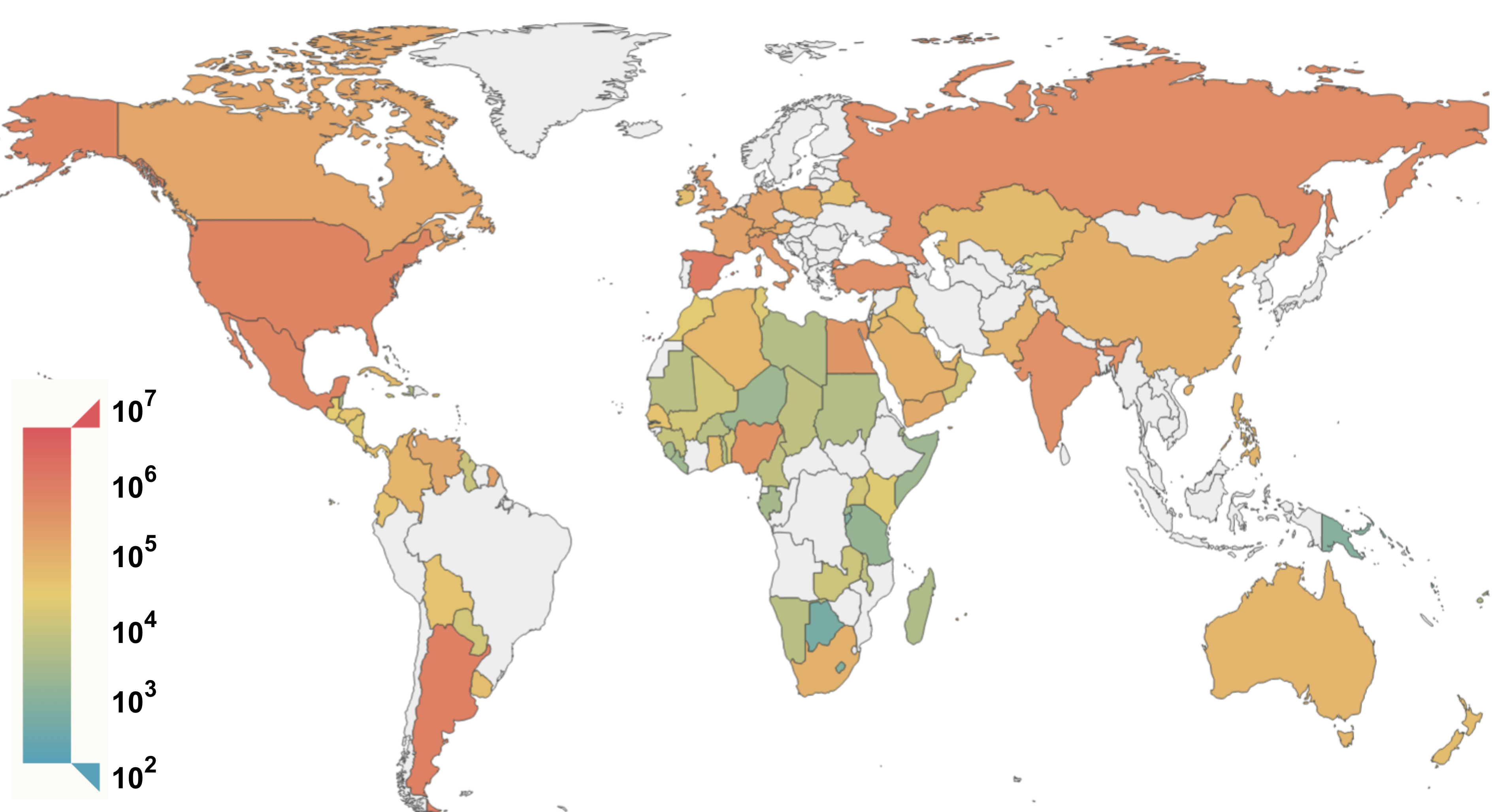} 
\caption{The number of news articles per country in our base dataset. The released dataset includes labels for pairs of articles sampled from the base dataset.}
\label{fig:global_map}
\end{figure}

\section*{Dataset Creation}
The news articles are collected from Media Cloud, an online platform that since 2009 has collected more than a billion news articles published globally \cite{Roberts2021a}. 
We focus on news articles in ten languages, used as an official language in 124 countries. This set of countries covers 64\% of all countries and about 76\% of the world population (Figure~\ref{fig:global_map}).
%
Overall, we collected metadata and full text of all news articles from January 1, 2020, to June 30, 2020, totaling  $\sim$60M news articles in the following languages: 
English, Spanish, Russian, German, French, Arabic, Italian, Turkish, Polish, and Mandarin Chinese.

We recruited and paid annotators with fluency in these languages, to label the news article similarity after rounds of training and calibration.
In order to highlight the news similarities that bear significant value for communication and social science research, we meticulously selected relevant news articles. From this curated pool, we then systematically sampled pairs of articles for detailed annotation (Figure~\ref{fig:data_pipe}). The annotated data is available on at \url{https://zenodo.org/records/10611923}.\footnote{\url{https://zenodo.org/records/10611923}}

\paragraph{News article selection.}

Collected metadata of news articles varies in completeness, and thus it cannot be processed in the same way. We omitted the articles that miss any of the following basic attributes: story ID, URL, title, and text; and only included the meaningful and informative news articles with at least 100 word count (after translating into English).\footnote{We apply this threshold after translating to English to make a fair comparison across languages since the average number of words needed to represent an English document in another language depends on that language.} Also if a news article has the exact same title or URL as another newer one, we filter it out as a duplicate. Finally, we took some websites out of consideration since they are not trustworthy information sources, or do not focus on societal and political topics \footnote {Irrelevant websites include: ``reddit.com,'' ``facebook.com,'' ``twitter.com,''  ``fb.com,'' ``wikipedia.org,'' ``epochtimes.com,'' ``youtube.com,'' ``slideshare.net''. Any URL containing ``sport'' is also dropped.}.

\begin{table*}[t]
    \centering
    \footnotesize
    \newcommand{\vsim}{VS}
    \newcommand{\ssim}{SS}
    \newcommand{\sdis}{SD}
    \newcommand{\vdis}{VD}
    \newcommand{\oth}{OT}

    \newcolumntype{L}[1]{>{\raggedright\let\newline\\\arraybackslash\hspace{0pt}}p{#1}}
    \begin{tabular}{*{2}{L{3.5cm}} *{6}{p{0.6cm}} *{6}{p{0.75cm}} *{1}{p{1.2cm}}}
        \toprule
         Article 1 & Article 2 & GEO & ENT & TIME & NAR  & STY & TONE & FRAME& OVERALL \\ 
        \midrule

        Key dates in Indiana's path to reopening amid the coronavirus pandemic
        & 
        Outlining “measured” lifting of lockdown
        & \vsim & \vsim & \vsim & \vsim & \sdis & \vsim & \textbf{\vsim} & \vsim\\[1cm] 
         
         Concor Head rallies SA to fight virus
         & 
         
         Church Leader Defies Coronavirus Measurers
         
         & \vsim & \sdis & \sdis & \sdis & \ssim & \vdis & \textbf{\vdis} & \sdis\\[0.5cm] 
         
         VDH confirms 1st coronavirus-related death in Virginia (March 14, 2020)
         & 
         Virginia sees new coronavirus cases (April 15, 2020)
         & \vsim & \ssim & \sdis & \ssim & \sdis & \ssim & \textbf{\ssim} & \ssim\\ 
        
        \bottomrule

    \end{tabular}

         
         
         
         
        
        
    \caption{Annotation example of news article pairs. Each aspect is labeled as one of the four classes: Very Similar (VS), Somewhat Similar (SS), Somewhat Dissimilar (SD), Very Dissimilar (VD). The first example includes a pair of articles that cover the plan for reopening Indiana, which are very similar in frame. The second pair shares some similarities in terms of Geography (South Africa) and Entities (President Cyril Ramaphosa), but their frames on the lockdown events exhibit significant dissimilarity. One article discusses how an infrastructure company supports the national lockdown for long-term commercial sustainability, while the other focuses on church leaders' disagreement with the lockdown due to its limitations on in-person worship. The final pair of articles both report on coronavirus cases in Virginia, which are somewhat similar in their framing. The first article includes mayoral statements and conveys a sorrowful tone, while the second article presents the information in a more neutral tone, focusing on data.}
    \label{tab:examples}
\end{table*}

After applying these selection measures, the resulting news article counts per language are as follows: English (10M articles), Spanish (4.6M), Russian (1.8M), German (1.3M), French (1.2M), Arabic (1.8M), Italian (1.5M), Turkish (655K), Polish (369K), and Mandarin Chinese (205K).

\paragraph{News article pairs sampling.}
Ideally, the dataset should cover each degree of news article similarity in an even manner. However, if we just sample two random news articles they can hardly be relevant, let alone similar. A crucial challenge of our work was identifying potentially similar pairs of news articles, in order to mitigate the imbalance of similarity labels during the annotation process. We experimented with various strategies to unearth these similar articles, including the comparison of document embeddings \citep[Cr5:][]{josifoski-wsdm2019-cr5} or sentence embeddings \citep[Sentence BERT:][]{reimers-gurevych-2019-sentence} for news headlines and leading paragraphs, as well as the named entities extracted from full texts \citep[Babelfy, polyglot, and spaCy:][]{moro2014entity}.


After extensive pilot study, including the selection of news article representation for sampling (e.g., we tried Cr5 and Sentence BERT embeddings, but concluded that Wikified named entities work better), filtering process, and developing simple machine learning models for active learning, we ultimately engineered an effective and efficient pipeline to sample and annotate pairs of news articles (Figure~\ref{fig:data_pipe}). Our first step is to extract named entities from each news article, a process facilitated by the use of spaCy and Polyglot. 


To avoid the sampling of duplicate pairs (i.e., two articles with identical or nearly identical text, published under different titles and URLs and hence not filtered out during the article selection stage), we introduced an additional filtering step. This process removed all pairs of articles that share one or more long sentences (comprising 40 or more characters) or whose Jaccard similarity of article text exceeds 0.25, a predetermined empirical threshold. 

As human annotators started to label the news article pair samples, the sampling quality was iteratively improved via a human-in-the-loop active learning framework. In essence, with each iteration, we built a fresh new logistic regression model and trained it with all the labels that we had obtained up to that point. The model includes features as follows: the word counts of both articles, the number of common words, the number of common named entities, cosine similarity of the named entities with BM25 embeddings \cite{bm25_Robertson_2009}, text Jaccard similarity, and an exponentially decaying function of publication date difference.




\subsection*{Annotation Process}

In consultation with media studies literature, we established an elaborate codebook, which we share with our dataset,\footnote{\url{https://zenodo.org/records/10611923}} that details the annotation of news article similarity across 8 aspects (Figure~\ref{fig:annotation-scheme}). To calibrate the annotators' understanding of the codebook, we initially conducted several rounds of trial annotations, including calibration sessions aiming for agreement on 30 carefully selected news article pairs. The subsequent annotations that contributed to the overall dataset were collected in two stages, described next. 

In the first stage (July - December 2021), annotators assessed the first seven similarity aspects (GEO, ENT, TIME, NAR, OVERALL, STYLE, TONE). During this time, many article pairs were assigned to multiple annotators to ensure label reliability. This stage resulted in 9,868 annotated pairs, which we released at the SemEval'22 competition focused on news article similarity estimation~\cite{chen2022semeval}.

\begin{table}[tbh]
    \centering

    \scalebox{0.95}{
    \begin{tabular}{crr}
    \toprule
    language class & count &mean(OVERALL)\\
    \midrule
    ar      & 1725 & 2.66  \\ 
    de      & 3402 & 2.54  \\
    en      & 7545 & 2.76 \\
    es      & 2334 &  2.33\\
    fr      & 369  & 1.99 \\
    it      & 1133 & 2.64 \\
    pl      & 760 & 2.33 \\
    ru      & 350 & 2.77 \\
    tr      & 1310 & 2.58 \\
    zh      & 3413 & 2.31\\
    de-en    & 1956 &2.94\\
    de-fr    & 178 & 1.92\\
    de-pl    &  47 & 1.69\\
    es-en    & 759 & 2.84\\
    es-it    & 474 & 2.29\\
    fr-en    & 103 & 1.21\\
    fr-pl    &  53 & 1.67\\
    pl-en    & 119& 2.35\\
    zh-en    & 525 & 2.96\\

    \midrule
    total & 26555 & 2.59\\
    \bottomrule
    \end{tabular}
    }
    
    \caption{Counts of annotated monolingual pairs (e.g., English article pairs -- "en") and cross-lingual pairs (e.g., pairs of a German article and an English article -- "de-en").}
    \label{tab:annotators}
\end{table}

Then the second stage followed (January - May 2022), during which the FRAME aspect was added. According to our codebook, FRAME similarity ``can be judged only when the framing and opinions communicated in the two articles target the same subject, e.g., the articles communicate opinions about Bernie Sanders''.
The idea that framing is relative to a target comes from the research area of targeted sentiment analysis.
Due to this requirement, and because we introduced FRAME aspect gradually while discussing it with our expert annotators, we collected FRAME similarity labels for a relatively small subset of 1522 news article pairs.\footnote{The vast majority of dissimilar news article pairs do not share common targets and thus their FRAME similarity is not labeled.}
Each pair was labeled by a single well-trained annotator to maximize the number of annotated news article pairs. Overall, the second stage brought 16,687 annotated pairs.

Each aspect was labeled with four ordinal classes -- \textit{Very Dissimilar}, \textit{Somewhat Dissimilar}, \textit{Somewhat Similar}, and \textit{Very Similar}. To accommodate exceptional cases, we also offer an \textit{Other} option. The most common exceptions that we met were pairs of duplicate new articles or inaccessible articles, for instance, due to paywalls or removal (annotators were instructed to report such instances via a free-text comment).


To satisfy the desired linguistic diversity and scale of news annotation, we trained 25 annotators, recruited from three institutions (GESIS, UMass, Umich). They did several rounds of calibration for the gold standards in our codebook \citet{chen2022semeval} before starting their annotation tasks. The entire annotation process lasted for roughly 11 months. We paid each annotator €12 per hour at GESIS and $\$15$ per hour at UMass and UMich.

The annotation process was running through a custom annotation interface devised by our team. It shuffles news article pairs and assign them to annotators as per their respective language capabilities. To engage and motivate the annotators, the interface also offers statistical feedback, such as the annotation count ranking and the inter-rater agreement ranking among annotators. This also enabled us to identify the annotators who did not perform their task correctly. We recorded the disagreements and discussed them with annotators biweekly as part of our iterative calibration process.

\section*{Dataset Description}

\subsection*{Format and Statistics}

We collected the labels for roughly 11 months from 2021 July to 2022 May. Ultimately, we got the news article similarities for 26,555 pairs (Table \ref{tab:examples}), including 1522 with frame similarity labels and 4,214 cross-lingual pairs (Table \ref{tab:annotators}). In the remainder of this manuscript, we represent the four similarity labels on an interval scale from 1 (Very Dissimilar) to 4 (Very Similar), e.g., Table \ref{tab:annotators} shows the average similarity for different groups of labels.

\subsection*{Inter-annotator Agreement}

To evaluate the reliability of the similarity labels, we calculated the inter-rater agreement for each aspect. The annotators exhibited remarkably high agreement on the OVERALL similarity aspect, as evidenced by a Krippendorff's~$\alpha$ of $0.77$. We also leveraged Gwet's AC$_{1}$ measure, which is known to be less sensitive to non-uniform marginal label distributions~\cite{gwet2008computing}. Therefore, it allows us to offset bias arising from skewed distributions within some aspects.  All aspects demonstrated good inter-rater agreements under this measure.


\section*{Applications}

To the best of our knowledge, our dataset is the largest to date for assessing news article similarity, with meticulous evaluations conducted across multiple languages. Consequently, it holds significant potential to empower a wide range of applications within the fields of media communication and social science, including news aggregation, media consumption analysis, cross-cultural studies, agenda setting research, linguistic studies, and political science research. Next, we present three examples (our code of implementations is available on Github.\footnote{\url{https://github.com/social-info-lab/global_news_synchrony}}.

\begin{table}
\Huge
    \centering

    \resizebox{0.48\textwidth}{!}{
    \begin{tabular}{rrrrrrrr}
    \toprule
     &  GEO &  ENT & TIME  & NAR & OVERALL & STYLE & TONE \\
    \midrule
Krip. $\alpha$  & 0.73 & 0.69 & 0.57 & 0.69 & 0.77 & 0.46 & 0.38 \\
Gwet's $AC_1$ & 0.75 & 0.70 & 0.78 & 0.71 & 0.79 & 0.60 & 0.60 \\
    \bottomrule
    \end{tabular}
    }
    
    \caption{Inter-rater agreement measures, Krippendorf's $\alpha$ and Gwet's AC$_{1}$, for the similarity aspects (FRAME aspect is missing since for each pair it is only annotated by one well-trained annotator).}
    \label{tab:agreements}
\end{table}

\begin{table*}[htb]
    \centering
    \scalebox{0.9}{
    \begin{tabular}{c|l}
    \toprule
    \textbf{Event}   &   \textbf{News article title} \\
    \midrule
    \multirow{10}{*}{Oscar 2020} & Oscary 2020 należały do "Parasite" i zmieniły historię gali. Pełna lista zwycięzców \\
    & Oscar 2020: revisa la lista de ganadores de las 24 categorías con lo mejor del cine \\
    & Oscars 2020: Los usuarios de internet premian a Joker, Leonardo DiCaprio, Scarlett Johansson y Martin Scorsese\\
    & A dos horas de la ceremonia, los Oscar se aprontan entre favoritos y posibles sorpresas\\
    & Parasite hizo historia en los Oscars y se llevÃ³ el premio a mejor pelÃ cula\\
    & Ganadores de los Oscar 2020 \\
    & Nigeria Records 245 New Cases Of COVID-19, Highest Single-Day Increase \\
    & Oscars 2020: South Korean movie makes history by winning best picture \\
    & Múltiples latinos se miden esta noche en los Oscar \\
    & Glamour y talento en la gala de los premios Oscar en su 92 edición \\
    & Oscar night begins with `1917' battling `Parasite' \\
    \midrule
    \multirow{11}{*}{Covid 2020} & Canadá acepta tener varios posibles casos de coronavirus \\
    & Prevén se presenten casos de coronavirus chino en México \\
    & Qué se sabe sobre el coronavirus de China que "puede haber afectado a cientos de personas", según científicos británicos \\
    & Chinese confirm coronavirus outbreak can spread like wildfire from infected people | Fox Business\\
    & China locks down more cities as virus spreads\\
    & US source: North Korean leader Kim Jong Un in grave danger after surgery\\
    & Un virus en Chine commence à inquiéter au Canada\\
    & El nuevo coronavirus deja 106 muertos y 4.515 infectados en China\\
    & Lo que se sabe sobre el coronavirus detectado en China y otros países que ya ha afectado a cientos de personas\\
    & WDH/VIRUS/ROUNDUP: Vorerst keine `internationale Notlage'\\
    & Virus chinois: sans doute des centaines de contaminations, inquiétude à l étranger\\
    \bottomrule
    \end{tabular}
    }
    \caption{Random samples of news articles in two exemplary news event clusters: Oscar 2020 covered by news in English and Spanish, with a small portion in Polish; Covid 2020 covered by news in various languages from Asia, North America, South America and Europe.}
    \label{tab:cluster_exp}
\end{table*}

\subsection*{Global News Synchrony and Diversity}
As the multilingual news article similarity offers a unified representation that transcends language barriers, it enables us to understand the news media across multiple countries, or even on a global scale. In a recent paper, we develop an effective methodology for news coverage studies at a massive scale and measure news diversity and synchrony across countries~\cite{xi2024globalnews}.

\textbf{Challenges.} The key challenges to examining news coverage at a global scale are the following. 
First, traditional data collection and validation based on physical newspapers and questionnaires requires human effort that scales linearly with the amount of data and the number of languages. 
These practical considerations severely limit the data size and linguistic coverage of traditional studies.
Second, it is not clear how to identify global news events, which are necessary to measure news coverage of events. 
Existing methods for identifying which events are reported in the news prioritize precision over coverage, since such methods are based on keyword matching \cite{card2015media}, inevitably leading to the lack of generality.
We overcome these challenges by contributing a novel computational methodology for studies of global news coverage thanks to the labeled dataset.

\textbf{News similarity inference.} 
First, we develop
a computationally-efficient transformer model that infers multilingual news similarity. The model achieves the highest score (Pearson correlation with human annotations of $0.8$) among similar efficient models in the prior benchmark~\cite{chen2022semeval} computed on our labeled dataset, achieving performance comparable to average human annotators.

\textbf{Global event detection.}
Second, using this model, we compute similarity among millions of news article pairs. Then, we apply a graph-clustering algorithm on the resulting similarity network to identify 4,357 multilingual news events.
The largest events, in chronological order, were: the assassination of Iranian general Soleimani, U.S. presidential election primaries, the Covid-19 pandemic, and protests after the killing of George Floyd (see examples in Table \ref{tab:cluster_exp}). We evaluate the quality of the identified news events by an intrusion task, which is commonly used to evaluate topic models~\cite{chang2009reading}. We recorded an average high precision of $85.8\%$ across the annotators ($97.5\%$ for the annotator who spent the most time on the task).

\textbf{News diversity and synchrony measures.}
Third, we introduce information-theoretic measures of country-level synchrony and diversity in news coverage of global events. We define the \textit{diversity} of news in a country as the entropy of the distribution of the news published in that country across the inferred events. As the \textit{synchrony} of news across a pair of countries we define the Jensen-Shannon divergence of the respective distributions. Next, we regress the diversity within a country and synchrony across countries against country-level predictors. The regression yields much higher adjusted $R^2$ for the introduced measures of diversity and synchrony ($R^2$ of 0.54 and 0.45, respectively) than naive baseline measures based on an average of pairwise news article similarity that do not make use of global news events ($R^2$ of 0.13 and 0.30, respectively).

\begin{figure*}[t]
\centering
\includegraphics[width=2\columnwidth]{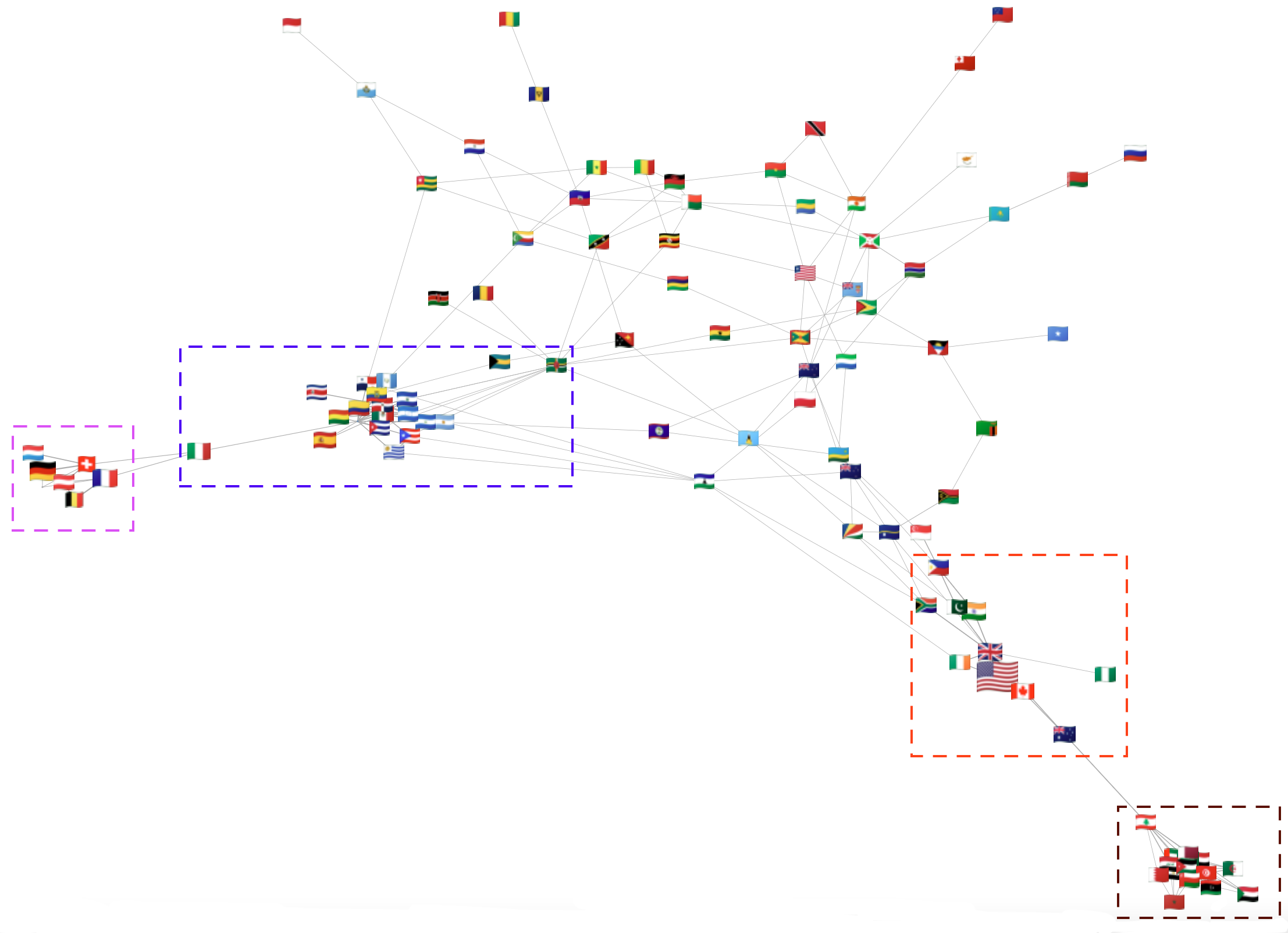} 
\caption{News event graph backbone for the top 100 countries with the largest populations. The main communities are marked with squares: the US and the UK and their former colonies (red square), five countries of the old European Union of 1958 (purple), Latin America and Spain (blue), and the Arab world (brown). Most of these communities align with the geographic-linguistic groups in \cite{kim1996determinants}.
Some countries are not selected into the graph backbone, e.g., China. Country flag size corresponds to country GDP, and each edge represents 95\% confidence~\cite{serrano2009extracting} that it represents non-random synchrony in news event coverage.}
\label{country_graph}
\end{figure*}

\textbf{Findings.} 
The labeled dataset and proposed methodology enable the discovery of unexpected patterns in global news coverage. 
For instance, prior studies suggest that the acceleration of the news cycle in the Internet age contributes to the homogenization of news coverage~\cite{bucy2014media, Boczkowski2007-ip, mcgregor2019social, zuckerman2013rewire}. However, we find that Internet adoption is the \textit{strongest} predictor of news diversity within a country ($p<0.005$). The higher the Internet penetration, the larger the news diversity, probably because news media cater to diverse interests and motivations of online audiences \cite{lee2013news}). 
This finding illustrates the potential of the proposed methodology to contribute to communication science in ways that go beyond confirmations of existing theories.
In addition, we find that news coverage is more diverse in countries with multiple official languages ($p<0.005$), more diverse religious practices ($p<0.005$), greater economic disparities, and larger populations ($p<0.05$). 

The international news synchrony network based on the proposed synchrony measure reveals groups of countries that synchronize in their news coverage of events (Figure~\ref{country_graph}): (i) the US, UK, and UK's past colonies, (ii) the old European Union of 1958, (iii) Latin America, and (iv) Arab countries. 
We find that trade volume is the strongest predictor of news synchrony between countries ($p<0.005$), which corroborates prior findings \cite{wu2000systemic,segev2016group}. 
Furthermore, coverage of news events is more synchronized between countries that share an official language ($p<0.005$), high GDP ($p<0.05$), and high democracy indices ($p<0.005$). 
Interestingly, countries that belong to NATO experience more news synchrony ($p<0.05$), possibly because they have common security concerns and some of the largest news events correspond to military operations. Countries belonging to BRICS ($p<0.05$) exhibit more synchronized news, possibly due to their common developmental interests.

\subsection*{Media Bias Analysis}\label{subsec:media_bias}

\begin{figure}[htb]
\centering
\includegraphics[width = 1.6in]{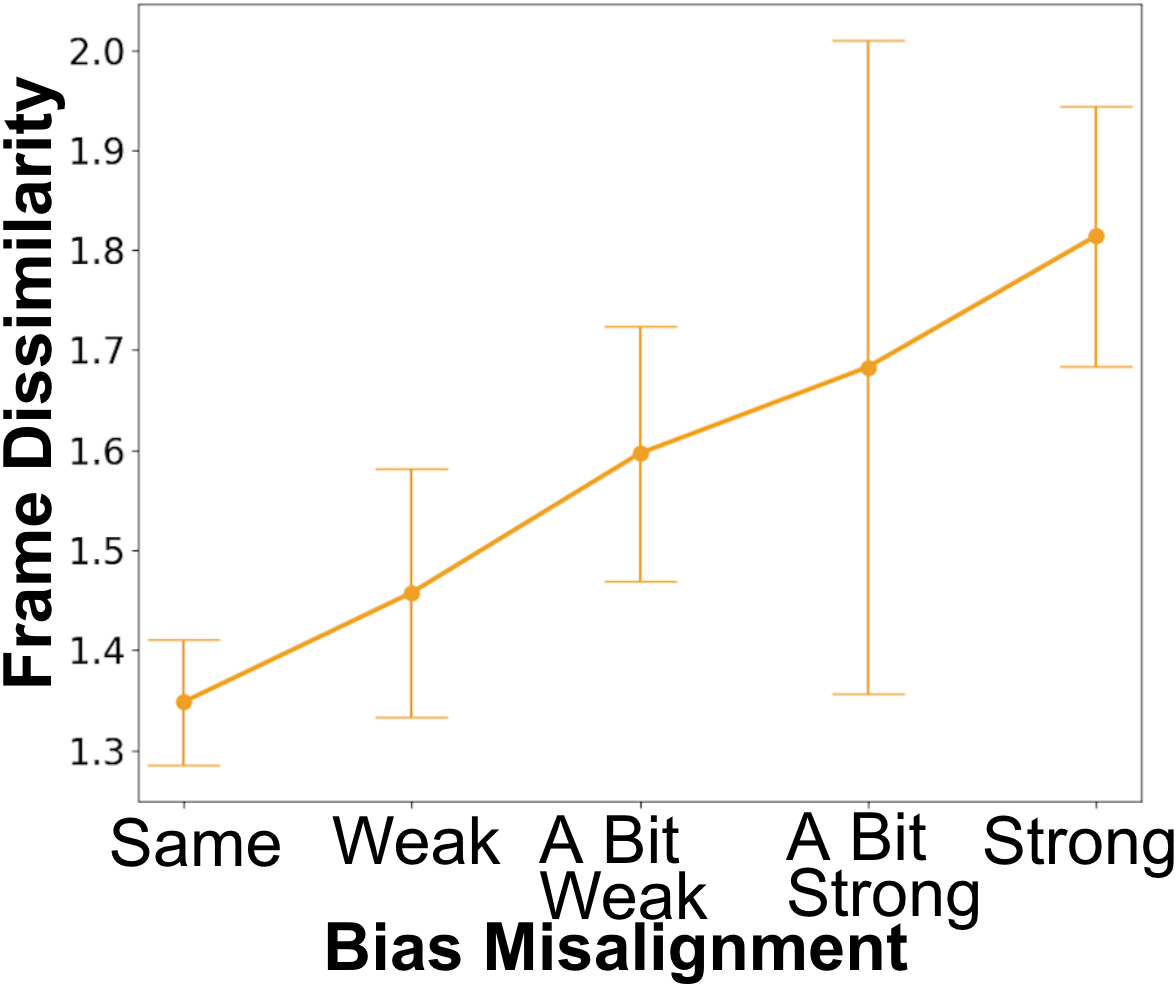}
\includegraphics[width = 1.62in]{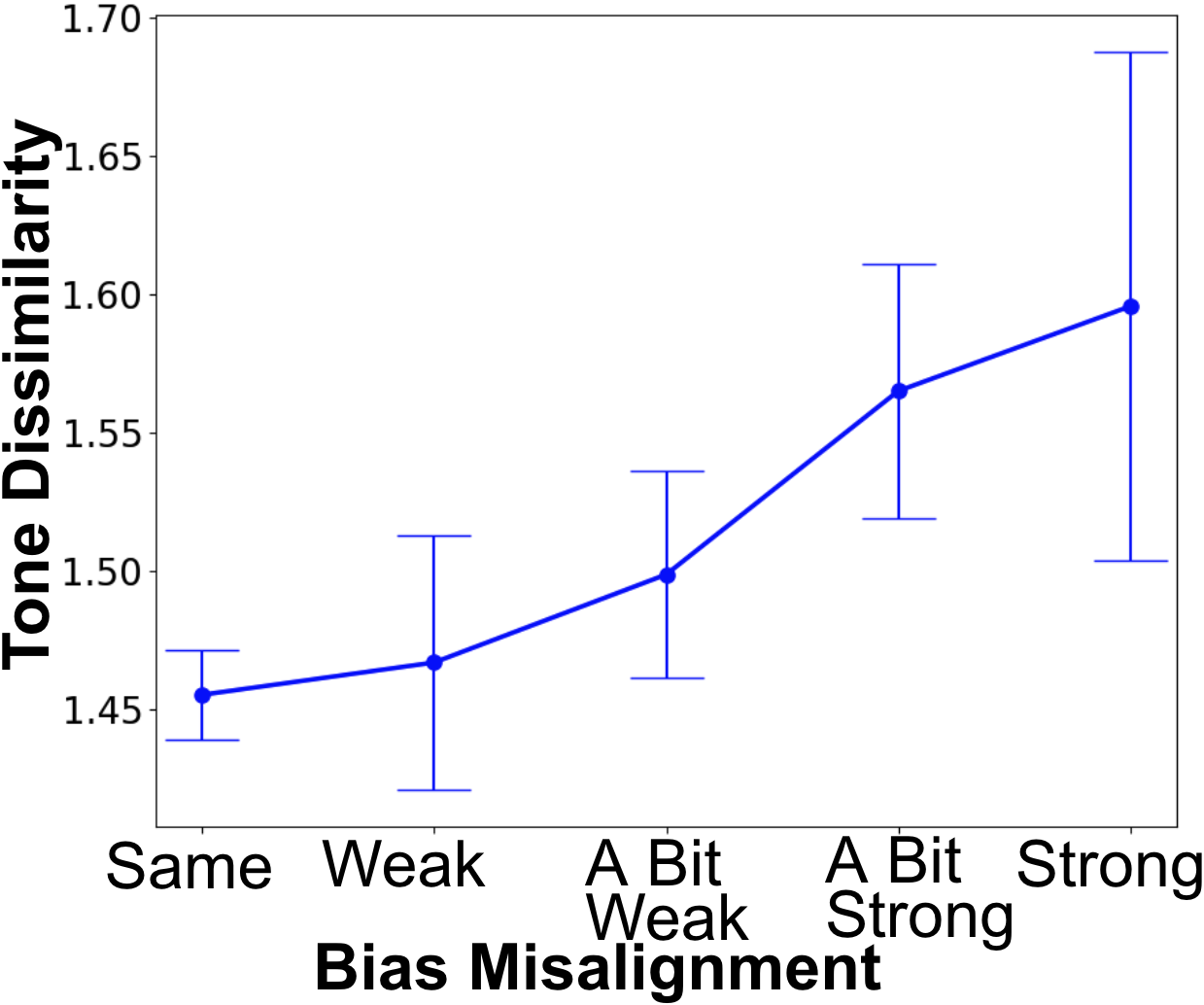}
\caption{The average frame dissimilarity (left) and tone dissimilarity (right) for pairs of media outlets with a certain level of political bias misalignment.}
\label{fig:frame_frame_tone_tone}
\end{figure}

Media outlets often exhibit biases in their coverage of events and the emphasis they place on them. These biases are influenced by various factors such as political stances, national interests, cultural beliefs, and target audiences \cite{mrogers1988agenda}. These social factors significantly affect how a story is presented, including its frame and tone \cite{aruguete2017agenda}. The framing of a story can direct audiences towards specific truths or opinions, especially those that resonate with their existing beliefs and knowledge \cite{scheufele2007framing}. Additionally, the tone of a story can subtly guide the conveyance of ideology from media to the public, leveraging audience conformity \cite{ambady2008first, lebon2017crowd}.

\textbf{Media bias vs news frame and tone.} Here, we study the relationship between media outlet bias and the framing and tone of the news articles they publish.

We compiled a list of media outlet biases from the Media Bias/Fact Check website \cite{MBFC2022}. Additionally, we consulted three supplementary sources of media bias scores: (1) a dataset on partisanship from the 2016 American presidential election, with biases inferred based on whether a Twitter user followed Hillary Clinton or Donald Trump \cite{faris2017partisanship}; (2) voter registration data for Democrats and Republicans from 2018 \cite{robertson2018auditing}; and (3) an analysis of political bias among active Twitter users from January 2019 to June 2019, utilizing the emIRT tool \cite{imai2015emirt, mediacloud2020pub}. These biases span five categories on the left-right spectrum: \textit{Left}, \textit{Left-center}, \textit{Center}, \textit{Right-center}, and \textit{Right}. 
As the bias of an outlet, we used the label that was the most common across the above four sources. In cases where this procedure identifies varying bias labels, we use the most left-leaning one.



We plot the frame and tone dissimilarity as a function of media bias misalignment (Figure~\ref{fig:frame_frame_tone_tone}),
classified into five categories based on the ideological distance on the spectrum from \textit{Left} bias to \textit{Right} bias: \textit{Same}, \textit{Weak}, \textit{A Bit Weak}, \textit{A Bit Strong}, and \textit{Strong}. 
For example: if one news outlet has a \textit{Left} bias and the other has a \textit{Right} bias, then the bias misalignment is \textit{Strong}; but if the second article has a \textit{Center} bias, then the bias misalignment is \textit{Weak}.
As expected, our findings reveal that the frame and tone of a pair of news articles become more dissimilar if the biases of the media outlets where they were published are more misaligned (Figure~\ref{fig:frame_frame_tone_tone}).


\begin{table}[htb]
    \centering

    \resizebox{0.48\textwidth}{!}{
    \begin{tabular}{cccccccc}
    \toprule
     & Frame vs Bias & Tone vs Bias & Frame vs Tone \\
    \midrule
Spearman & 0.185 & 0.087 & 0.394\\
Pearson &  0.202 & 0.083 & 0.410\\
    \bottomrule
    \end{tabular}
    }
    
    \caption{Correlations between frame similarity, tone similarity, and media bias. All the correlations are statistically significant with p-values less than $10^{-9}$. }
    \label{tab:syn_frame_tone_politics}
\end{table}

We also calculated the correlations between frame dissimilarity, tone
dissimilarity, and media bias misalignment at the level of article pairs (Table \ref{tab:syn_frame_tone_politics}). We find a modest correlation between frame/tone similarity and media bias alignment. Notably, frame similarity demonstrates a slightly stronger correlation than tone similarity, although both correlations are relatively weak. The correlations between frame and tone are both around 0.4, which indicates medium strength.

\begin{figure}[htb]
\centering
\includegraphics[width = 2.2in]{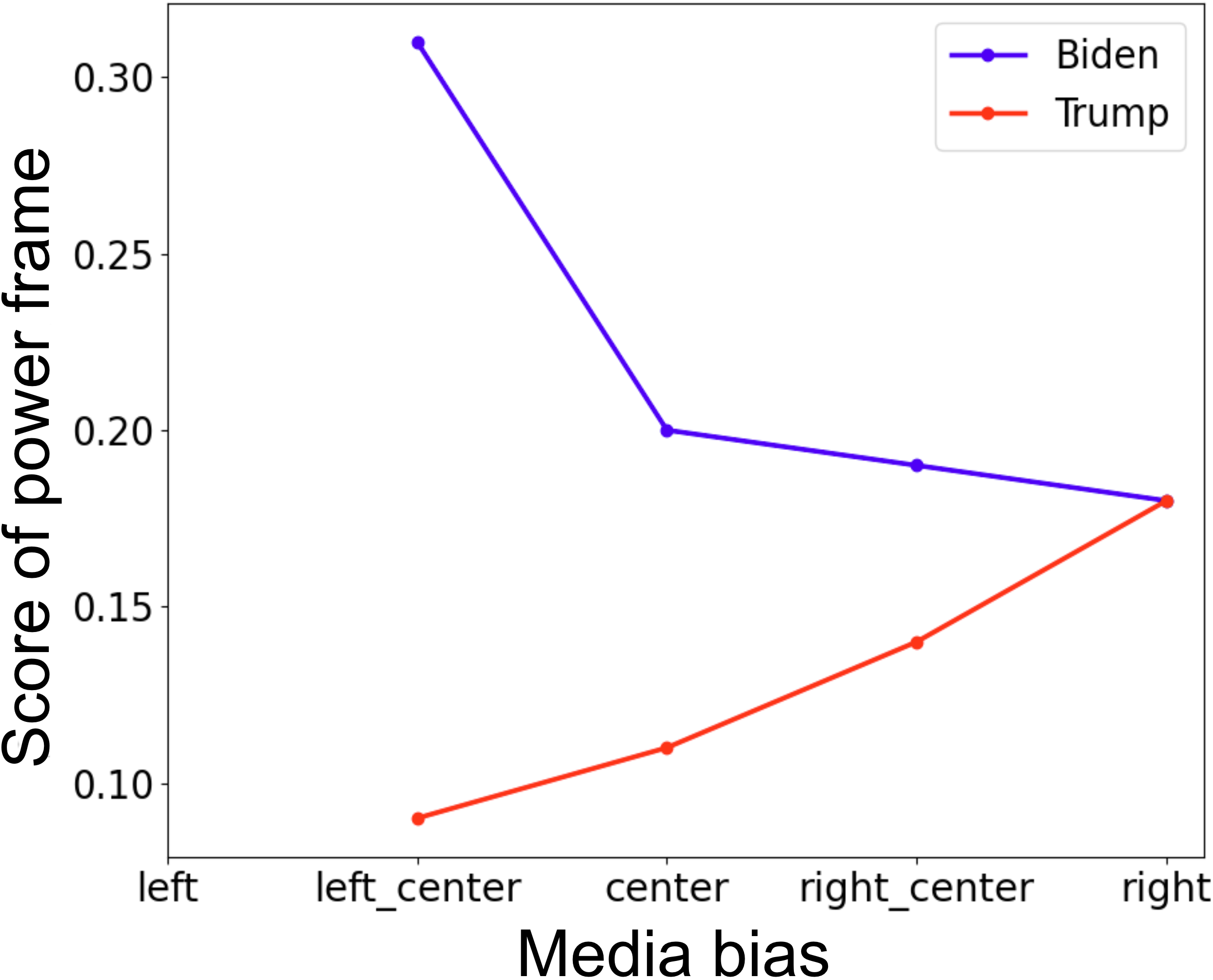}
\caption{Power scores of Biden and Trump inferred by Riveter based on news articles published in news outlets with given media bias.}
\label{fig:bias_politician}
\end{figure}

\textbf{Media bias on president power.} 
News outlets may portray biased social images of politicians. To investigate this phenomenon, we utilized Riveter \cite{antoniak-etal-2023-riveter}, an advanced tool capable of identifying named entities within each article and assigning a power score to each of these entities. This power score reflects the entity's perceived power strength, based on the actions associated with it in the connotation frame \cite{sap-etal-2017-connotation}. Our process began with applying Riveter to deduce the power scores of named entities. We then merged the coreferences of these entities and excluded those appearing in articles from fewer than three different news outlets. Subsequently, we segmented the remaining articles into categories based on differing political biases. Our analysis revealed that across all categories, news outlets consistently depicted Biden as having higher power strength than Trump (Figure~\ref{fig:bias_politician}). This could be attributed to their inherent social images in public perception (such as Biden usually refers to collaborative power words like U.S. institutions, achievements, and morality, consistent with the constructs of prestige and traditional power; while Trump is around coercive power words like “defeat,” “threat,” “poison,” “administration,” “failed,” “ignored,” or “promised.”, which may be interpreted as denial of Trump’s trustworthiness and dependability \cite{korner2022linguistic}. Interestingly, outlets with a bias leaning towards the \textit{Left} portrayed Biden as more influential than those leaning towards the \textit{Right}, with the opposite trend observed for Trump, similar as the political biases of parties they two respectively represent and obtain support from (Democratic for \textit{Left} and Republic for \textit{Right}).



\subsection*{Multi-factor Analysis}

Our dataset allows for a nuanced analysis that considers multiple factors together and thus extends beyond a single social factor to analyze how various elements like political biases, language use, and country-specific factors interact to influence news similarity. By integrating these diverse aspects into our investigation, we provide a more complete picture of the dynamics shaping global news narratives, offering deeper insights into how political, cultural, and national factors collectively influence news coverage and framing.

Next, we quantify the importance of each of these factors on OVERALL, TONE, and FRAME similarity of news articles. To this end, we built three logistic regression models and compared the factors' coefficients with each other. In these models, the pairwise alignment of each factor becomes a feature and they are combined as the input, while the news similarity of OVERALL, TONE, or FRAME aspect becomes output for each of the three classifiers.
For simplicity, we represent language alignment and country alignments as binary: 1 for exact matches and 0 for differences. Bias alignment is normalized to [0,1] range as ordinal factor with equal numeric interval ("Same" is converted to 1 and "Very Strong" is converted to 0), for comparing its importance with other factors at the same scale. The similarities are also recast into binary: 1 for \textit{Very Similar} and \textit{Somewhat Similar}, and 0 for \textit{Somewhat Dissimilar} and \textit{Very Similar}.

Table \ref{tab:syn_logit_perf} displays the logistic regression weights of each factor. Interestingly, bias alignment positively correlates to content similarity, possibly because contentious topics, often covered by politically polarized outlets, drive more discussion. Shared languages and shared countries publishing the news articles also showed a positive correlation to OVERALL similarity, reflecting common cultural beliefs and ties of national interest. However, the alignment of the countries mentioned in the news articles seems to correlate negatively with content similarity.

\begin{table*}[t]
\begin{center}
\begin{tabular}{l rl rl rl}
\toprule
\multicolumn{1}{c}{} & \multicolumn{2}{c}{\textbf{OVERALL}}& \multicolumn{2}{c}{\textbf{FRAME}} & \multicolumn{2}{c}{\textbf{TONE}}\\
\cmidrule(lr){2-3} \cmidrule(lr){4-5} \cmidrule(lr){6-7}
\textbf{Alignment} & \textbf{coef} & \textbf{P$> |$t$|$} & \textbf{coef} & \textbf{P$> |$t$|$} & \textbf{coef} & \textbf{P$> |$t$|$}\\
\cmidrule(lr){1-1} \cmidrule(lr){2-2} \cmidrule(lr){3-3} \cmidrule(lr){4-4} \cmidrule(lr){5-5} \cmidrule(lr){6-6} \cmidrule(lr){7-7}
{Bias}            &       -0.184  &  \textbf{**}  &  0.137  & \textbf{*}  & 0.091 & \textbf{**}\\
{Language}             &      0.097  &  \textbf{**} & -0.013  &  & 0.084 & \textbf{**}\\
{Publication country}       &      0.203  &   \textbf{*} &       0.128  & \textbf{*}  & 0.037 & \textbf{*}\\
{Referred country}       &      -0.248  &   \textbf{**} &    0.059  &   & 0.046 & \textbf{*}\\
\midrule
F1 Score & 0.515 && 0.689 &&  0.656 &\\
Accuracy & 0.530 && 0.638 &&  0.610 & \\
\bottomrule
\end{tabular}
\end{center}
\caption{Classification evaluation measures (F1, accuracy) and the coefficients of various predictors for the logistic regression models of OVERALL, FRAME, and TONE similarity. Significance level: difference from zero of more than one ($^*$) or three ($^{**}$) standard deviations, which correspond to p-values of less than $0.1$ and $0.001$, respectively.}
\label{tab:syn_logit_perf}
\end{table*}

It is worth noting that the strongest predictor of frame similarity is the bias alignment. The alignment of the country of publication also has some importance, potentially reflecting the influence of national interest on the news frame. In terms of predicting tone similarity, both bias alignment and language alignment demonstrate greater importance than country alignment. 
Table \ref{tab:syn_logit_perf} also demonstrates that frame and tone similarities are more predictable than overall content similarity. This observation aligns with the closer proximity of frame and tone to the chosen factors.



\section*{Ethics Statement}

Our similarity labels are based on full texts of news articles, but we only provide their URLs to access the full texts to prevent copyright issues. To ensure that the news content is reliable with a focus on socially meaningful topics or events, we declined all URLs from popular social media platforms (twitter.com, facebook.com, reddit.com, etc.). We encouraged the annotators to report such cases and introduced a button in our annotation interface to tag any content that they felt was hateful or harmful.


\section*{Conclusion}

In the ever-evolving landscape of global communication, understanding the interconnections between news articles is more than an academic pursuit—it's a key to unlocking insights into media studies and societal dynamics. Our extension of a multilingual news article similarity dataset sheds light on this intricate web by revealing commonalities across news articles in eight distinct aspects, including a novel approach to define news frames. This dataset is not just a tool; it's a window into the global media narrative, offering a unique vantage point for identifying international media networks, uncovering inherent biases in news outlets, and understanding the portrayal of presidential power across various media platforms.

However, the potential of this dataset extends far beyond these initial applications. It sets the stage for innovative research in global agenda setting, allowing for an in-depth exploration of media biases on a global scale. Imagine unearthing the stark disparities in news coverage, such as why African disasters require far more casualties to gain the same level of US media attention as those in Eastern Europe \cite{eisensee2007news}. Our dataset provides the granularity needed to dissect political campaigns and societal beliefs, revealing the subtle nuances that shape public opinion and discourse.

Moreover, the dataset is poised to facilitate critical analyses of long-term bias and synchrony trends in international news, e.g., in the periods leading up to war outbreaks. This kind of research promises to enhance our understanding of how media narratives intertwine with public sentiment, political maneuvers, international relations, and the genesis of global conflicts. In essence, it offers a lens to view and interpret the complex interplay of factors that drive the world's news stories.

By providing methodologies to observe and interpret the continuous evolution of our global narrative through the news, this work not only contributes to academic discourse but also offers profound societal insights. It underscores how international news coverage, in all its complexity, reflects and shapes our understanding of geopolitical histories and local realities. In doing so, this dataset stands as a pivotal resource for those seeking to comprehend the narrative of our global society.

\section*{Acknowledgments}
This research has received funding through grants from the Volkswagen Foundation and the University of Massachusetts Amherst. We sincerely thank Media Cloud for data access and the annotators for annotating multilingual news similarity. 

\bibliography{refs,aaai22}

\section*{Paper Checklist}

\begin{enumerate}

\item For most authors...
\begin{enumerate}
    \item  Would answering this research question advance science without violating social contracts, such as violating privacy norms, perpetuating unfair profiling, exacerbating the socio-economic divide, or implying disrespect to societies or cultures? \textcolor{red}{Yes.}
  \item Do your main claims in the abstract and introduction accurately reflect the paper's contributions and scope?
    \textcolor{red}{Yes.}
   \item Do you clarify how the proposed methodological approach is appropriate for the claims made? 
    \textcolor{red}{Yes.}
   \item Do you clarify what are possible artifacts in the data used, given population-specific distributions?
    \textcolor{gray}{NA.}
  \item Did you describe the limitations of your work?
    \textcolor{red}{Yes.}
  \item Did you discuss any potential negative societal impacts of your work?
    \textcolor{bluegreen}{No, because we haven't found any potential negative societal impacts.}
      \item Did you discuss any potential misuse of your work? \textcolor{gray}{NA.}
    \item Did you describe steps taken to prevent or mitigate potential negative outcomes of the research, such as data and model documentation, data anonymization, responsible release, access control, and the reproducibility of findings?
    \textcolor{red}{Yes.}
  \item Have you read the ethics review guidelines and ensured that your paper conforms to them?
    \textcolor{red}{Yes.}
\end{enumerate}

\item Additionally, if your study involves hypotheses testing...
\begin{enumerate}
  \item Did you clearly state the assumptions underlying all theoretical results?
    \textcolor{gray}{NA.}
  \item Have you provided justifications for all theoretical results?
    \textcolor{gray}{NA.}
  \item Did you discuss competing hypotheses or theories that might challenge or complement your theoretical results?
    \textcolor{gray}{NA.}
  \item Have you considered alternative mechanisms or explanations that might account for the same outcomes observed in your study?
    \textcolor{red}{Yes.}
  \item Did you address potential biases or limitations in your theoretical framework?
    \textcolor{gray}{NA.}
  \item Have you related your theoretical results to the existing literature in social science?
    \textcolor{red}{Yes.}
  \item Did you discuss the implications of your theoretical results for policy, practice, or further research in the social science domain?
    \textcolor{red}{Yes.}
\end{enumerate}

\item Additionally, if you are including theoretical proofs...
\begin{enumerate}
  \item Did you state the full set of assumptions of all theoretical results?
    \textcolor{gray}{NA.}
	\item Did you include complete proofs of all theoretical results?
    \textcolor{gray}{NA.}
\end{enumerate}

\item Additionally, if you ran machine learning experiments...
\begin{enumerate}
  \item Did you include the code, data, and instructions needed to reproduce the main experimental results (either in the supplemental material or as a URL)?
    \textcolor{red}{Yes.}
  \item Did you specify all the training details (e.g., data splits, hyperparameters, how they were chosen)?
    \textcolor{bluegreen}{No. Because this is dataset paper and its application details are not the focus.}
     \item Did you report error bars (e.g., with respect to the random seed after running experiments multiple times)?
    \textcolor{red}{Yes.}
	\item Did you include the total amount of compute and the type of resources used (e.g., type of GPUs, internal cluster, or cloud provider)?
    \textcolor{gray}{NA.}
     \item Do you justify how the proposed evaluation is sufficient and appropriate to the claims made? 
    \textcolor{red}{Yes.}
     \item Do you discuss what is ``the cost`` of misclassification and fault (in)tolerance?
    \textcolor{red}{Yes.}
  
\end{enumerate}

\item Additionally, if you are using existing assets (e.g., code, data, models) or curating/releasing new assets, \textbf{without compromising anonymity}...
\begin{enumerate}
  \item If your work uses existing assets, did you cite the creators?
    \textcolor{red}{Yes.}
  \item Did you mention the license of the assets?
    \textcolor{gray}{NA.}
  \item Did you include any new assets in the supplemental material or as a URL?
    \textcolor{red}{Yes.}
  \item Did you discuss whether and how consent was obtained from people whose data you're using/curating?
    \textcolor{gray}{NA.}
  \item Did you discuss whether the data you are using/curating contains personally identifiable information or offensive content?
    \textcolor{red}{Yes.}
\item If you are curating or releasing new datasets, did you discuss how you intend to make your datasets FAIR (see \citet{fair})?
\textcolor{red}{Yes.}
\item If you are curating or releasing new datasets, did you create a Datasheet for the Dataset (see \citet{gebru2021datasheets})? 
\textcolor{red}{Yes.}
\end{enumerate}

\item Additionally, if you used crowdsourcing or conducted research with human subjects, \textbf{without compromising anonymity}...
\begin{enumerate}
  \item Did you include the full text of instructions given to participants and screenshots?
    \textcolor{red}{Yes.}
  \item Did you describe any potential participant risks, with mentions of Institutional Review Board (IRB) approvals?
    \textcolor{red}{Yes.}
  \item Did you include the estimated hourly wage paid to participants and the total amount spent on participant compensation?
    \textcolor{red}{Yes.}
   \item Did you discuss how data is stored, shared, and deidentified?
   \textcolor{red}{Yes.}
\end{enumerate}

\end{enumerate}


\end{document}